\documentclass[letterpaper, 10 pt, conference]{ieeeconf}  % Comment this line out if you need a4paper

\IEEEoverridecommandlockouts                              % This command is only needed if 
\usepackage{graphicx} % for pdf, bitmapped graphics files
\usepackage{amsmath} % assumes amsmath package installed
\usepackage{amssymb}  % assumes amsmath package installed
\usepackage{makecell}
\usepackage{hhline}
\usepackage{rotating}
\usepackage{array}
\usepackage{multirow}
\usepackage{balance}

\usepackage{eso-pic}
\usepackage{hyperref}

\AddToShipoutPictureBG*{%
  \AtPageUpperLeft{%
    \hspace{0.5\paperwidth}%
    \raisebox{-1.2cm}{%
      \makebox[0pt][c]{%
        \footnotesize
        \begin{tabular}{c}
          Accepted to 2025 IEEE International Conference on Robotics and Automation (ICRA), Atlanta, USA \\
          DOI: \href{https://doi.org/10.1109/ICRA55743.2025.11128384}{10.1109/ICRA55743.2025.11128384}
        \end{tabular}
      }%
    }%
  }%
  \AtPageLowerLeft{%
    \hspace{0.5\paperwidth}%
    \raisebox{1.0cm}{%
      \makebox[0pt][c]{%
        \parbox{0.85\paperwidth}{%
          \centering
          \scriptsize
          \copyright\ 2025 IEEE. Personal use of this material is permitted. Permission from IEEE must be obtained for all other uses, in any current or future media, including reprinting/republishing this material for advertising or promotional purposes, creating new collective works, for resale or redistribution to servers or lists, or reuse of any copyrighted component of this work in other works.
        }%
      }%
    }%
  }%
}

\title{\LARGE \bf
Monotone Subsystem Decomposition for\\Efficient Multi-Objective Robot Design}

\author{Andrew Wilhelm$^{1}$$^{*}$ and Nils Napp$^{1}$% <-this % stops a space
\thanks{$^{1}$Department of Electrical and Computer Engineering, Cornell University,
Ithaca, NY, 14850, United States}%
\thanks{$^{*}$Correspondence to 
        {\tt\small ajw343@cornell.edu}}
\thanks{This material is based on work supported by the National Science Foundation grants NSF\#1846340, NSF\#2054744, and the GRFP DGE\#2139899. Any opinions, findings, and conclusions or recommendations expressed in this material are those of the author(s) and do not necessarily reflect the views of the National Science Foundation.}%
}

\begin{document}
\newtheorem{theorem}{Theorem}

\maketitle
\pagestyle{empty}
% \pagestyle{plain} %NN I like seeing the page numbers when working with printed PDFs

%%%%%%%%%%%%%%%%%%%%%%%%%%%%%%%%%%%%%%%%%%%%%%%%%%%%%%%%%%%%%%%%%%%%%%%%%%%%%%%%
\begin{abstract}
Automating design minimizes errors, accelerates the design process, and reduces cost.
% Automating design minimizes errors and accelerates the design process, thus reducing cost.
However, automating robot design is challenging due to recursive constraints, multiple design objectives, and cross-domain design complexity possibly spanning multiple abstraction layers.
Here we look at the problem of component selection, a combinatorial optimization problem in which a designer, given a robot model, must select compatible components from an extensive catalog. The goal is to satisfy high-level task specifications while optimally balancing trade-offs between competing design objectives.
% Here we look at the problem of component selection, a combinatorial optimization problem where a designer has a robot model and needs to select compatible components from a large catalog to meet high-level task specifications and optimally trade off between competing design objectives.
In this paper, we extend our previous constraint programming approach to multi-objective design problems and propose the novel technique of monotone subsystem decomposition to efficiently compute a Pareto front of solutions for large-scale problems. 
We prove that subsystems can be optimized for their Pareto fronts and, under certain conditions, these results can be used to determine a globally optimal Pareto front.
Furthermore, subsystems serve as an intuitive design abstraction and can be reused across various design problems.
% Furthermore, these subsystems serve as an intuitive design abstraction for engineers and can be reused across various design problems.
Using an example quadcopter design problem, we compare our method to a linear programming approach and demonstrate our method scales better for large catalogs, solving a multi-objective problem of 10\textsuperscript{25} component combinations in seconds.
We then expand the original problem and solve a task-oriented, multi-objective design problem to build a fleet of quadcopters to deliver packages. We compute a Pareto front of solutions in seconds where each solution contains an optimal component-level design and an optimal package delivery schedule for each quadcopter.

\end{abstract}

%%%%%%%%%%%%%%%%%%%%%%%%%%%%%%%%%%%%%%%%%%%%%%%%%%%%%%%%%%%%%%%%%%%%%%%%%%%%%%%%

% \vspace{-2.5mm}
\section{INTRODUCTION}
\vspace{-1.25mm}
\noindent 
Designing robots, especially optimal ones, is challenging for several reasons.
For instance, the search space of possible robot designs is large and can be both continuous and discrete. 
Moreover, many physics-based robot models have recursive constraints that need to be solved iteratively at each time-step and make optimization difficult. 
Finally, many approaches are heuristics-based and provide no guarantees of optimality of the final design. 
% Designing robots, especially optimal ones, is challenging for several reasons. 
% For instance, the search space of possible robot designs is large and can be both continuous and discrete. 
% Moreover, many physics-based robot models have recursive constraints that need to be solved iteratively at each time-step and make optimization difficult. 
% Finally, many approaches are heuristics-based and provide no guarantees of optimality of the final design. 

In automated robot design, an often overlooked challenge is the problem of component selection. 
During this stage, a designer must select components for a predefined robot model to meet high-level task specifications while optimizing one or more objectives.
% has already developed a robot model and must select components to use within the design to achieve high-level task specifications while minimizing or maximizing one or more objectives. 
% In considering the process of automated robot design, an often overlooked challenge is the problem of component selection, in which a designer has already developed a robot model and must select components to use within the design to achieve high-level task specifications while minimizing or maximizing one or more objectives. 
This is a combinatorial optimization problem that remains NP-hard even when formulated as an integer programming problem~\cite{schrijver_theory_2011}.
% This is a combinatorial optimization problem, and even if one casts the design problem as an integer programming problem,
% the problem is still NP-hard~\cite{schrijver_theory_2011}. 
% (e.g. selecting optimal components from a finisfste catalog to design a specific robot)
If one considers that parts suppliers such as DigiKey or Mouser\footnote{\texttt{www.digikey.com} and \texttt{www.mouser.com}} have catalogs that include tens of thousands of components, and designs are often optimized for several objectives, it becomes apparent that any approach must use sophisticated techniques to keep the problem tractable.

\begin{figure}[t]
      \centering
      
      \includegraphics[scale=0.32]{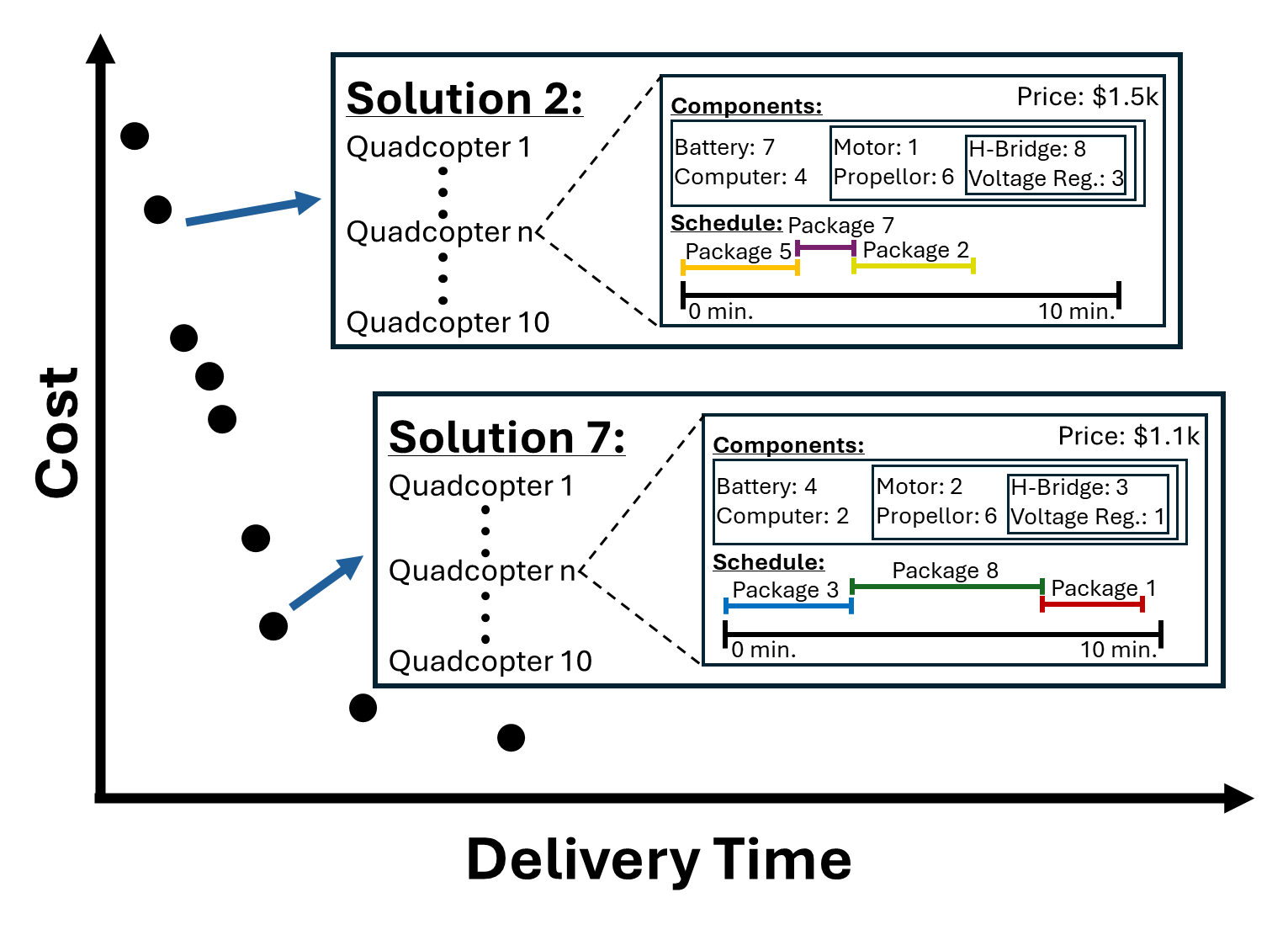}
      \vspace{-5mm}
      
      \caption[Caption for LOF]{\small We use our monotone subsystem decomposition technique to solve a large scale, multi-objective design problem where a fleet of quadcopters must deliver packages. By optimizing subsystems for individual Pareto fronts (PFs) and using these results to optimize the larger system, we can efficiently compute a globally optimal PF of solutions in seconds, where each solution on the PF provides the component-level design for each quadcopter and an individual delivery schedule for the packages that quadcopter has been assigned.}

      \label{overviewFig}
    \vspace{-5mm}
  \end{figure}

In previous work~\cite{wilhelm_constraint_2023}, we presented a framework using constraint programming (CP) for the single-objective component selection problem. In this paper, we extend this framework to multi-objective robot design, which is notoriously difficult due to the large search space. 
To address this, we introduce a new way to re-use optimization results of subsystems and still guarantee global optimality. 
% To address this, we introduce a novel idea to optimize subsystems within the larger problem, and we prove that using the resulting Pareto front of solutions from these optimized subsystems still results in globally optimal solutions to the larger problem. 
This technique speeds up computation by solving smaller problems that have smaller search spaces; saves future computation as optimized subsystems can be reused; and provides a designer useful insight into the optimal design of internal subsystems.

% Our approach is inspired by monotone co-design problems (MCDPs), which have nice abstraction and intuitive modeling for functionalities and resources. However, MCDPs require monotonicity as they use Kleene's algorithm to determine the Pareto front. Our approach uses monotonicity to ensure local optimization results produce globally optimal results, but our approach remains flexible in that it otherwise does not require monotonicity throughout the system and not even necessarily  in the subsystems that are being optimized. 

% The previous state-of-the-art approaches have used linear and integer programming
We compare our CP subsystem approach to a linear programming (LP) method from~\cite{carlone_robot_2019} using a quadcopter design problem.
Using component catalogs that are several orders of magnitude larger than previous related work, we demonstrate the scaling efficiency of our approach which can solve a multi-objective problem of 10\textsuperscript{25} candidate solutions and produce a Pareto front (PF) of optimal solutions faster than the LP approach can solve the same single-objective problem.

We further showcase the advantages of our approach by demonstrating its use in a large, complex case study: expanding the previous quadcopter design problem to 
design a quadcopter fleet for a multi-agent package delivery task, minimizing both delivery time and cost.
We can produce a PF of solutions (each consisting of a set of optimal, component-level quadcopter designs and a set of optimal schedules for each quadcopter) in seconds, see Figure~\ref{overviewFig}.

% We then further highlight the advantages of our CP subsystem approach by demonstrating its use in a large, complex case study. We expand the previous quadcopter design problem to produce a fleet of quadcopters for a multi-agent package delivery task, using as input high-level task specifications and producing designs across four levels of abstraction down to the component level. One of the advantages of using CP is the ability to interface with other common CP problems such as scheduling or packing, so we simultaneously solve a multi-objective scheduling problem to assign drones to packages for delivery in the shortest time while minimizing cost. 
% We can produce a Pareto front of solutions (each of which consisting of a set of optimal, component-level quadcopter designs and a set of optimal schedules for each quadcopter) candidate solutions in seconds.
% We can produce a Pareto front of solutions (each of which consisting of a set of optimal, component-level quadcopter designs and a set of optimal schedules for each quadcopter) from over 10\textsuperscript{\#} candidate solutions in fewer than \# minutes.

Our primary contributions are as follows:
\begin{itemize}
%    \item Developing subsystem decomposition for use with constraint programming to achieve efficient computing in larger design problems, and proving that this still provides globally optimal results under certain consistency conditions
%NN Version
        \item A decomposition technique for constrained optimization problems which allows reusing the optimization results of subsystems under certain consistency conditions
    % \item These subsystems are intuitive for designers and can provide insights into subsystem performance 
% \item Using this subsystem decomposition to solve a large design problem across four levels of abstraction. Our constraint programming approach unifies the problem of component selection and task planning into a single problem, allowing us to simultaneously solve a scheduling problem for a fleet of quadcopters and provide an optimal set of component designs to achieve this task.
\item Using this subsystem decomposition to solve a combined component selection and task planning problem across four levels of abstraction
% \item  This search space for this problem contains over XX possible solutions for just the quadcopter designs (not counting the complexity of continuous-time scheduling), orders of magnitude larger than previous research.
 
\end{itemize}

% The rest of this paper is structured as follows:
In the remainder of this paper, we first summarize related work (Sec.~\ref{sec:related-work}) and introduce notation for single- and multi-objective constrained optimization problems (COPs) (Sec.~\ref{sec:problem_background}).
% First we provide a brief background on related works (Sec.~\ref{sec:related-work}, and notation for single- and multi-objective constrained optimization problems (COPs) (Sec.~\ref{sec:problem_background}). 
We then provide the foundational framework for describing monotone subsystem decomposition (Sec.~\ref{sec:problem_formulation}). 
% We then discuss the foundational framework for applying monotone subsystem decomposition to component selection (Sec.~\ref{sec:problem_formulation}). 
% We then describe our framework for applying constrained optimization to component selection (Sec.~\ref{sec:problem_formulation}). 
In Sec.~\ref{sec:theoretical_results} we prove under certain conditions that the PF of a consistent subsystem can be used to determine a PF of globally optimal solutions, which is the theoretical basis for monotone subsystem decomposition.
Sec.\ref{sec:discussion} highlights advantages, discusses practical considerations, and contrasts our method with MCDPs, highlighting our method's greater flexibility. 
Finally, computational experiments validating our approach are presented in Sec.~\ref{computational_experiments}.
% In Sec.~\ref{sec:problem_formulation}, we then discuss the framework for applying constrained optimization to the problem of component selection. 
% Afterwards, in Sec.~\ref{sec:theoretical_results} we prove under certain conditions that the PF of a consistent subsystem can be used to determine a PF of globally optimal solutions.
% We then discuss these theoretical results in Sec.~\ref{sec:discussion}, discussing the advantages and practical considerations for using subsystem decomposition, addressing non-monotone systems, and comparing our approach to MCDPs and discussing how MCDPs impose different modeling constraints, with our monotone subsystem decomposition offering greater flexibility in certain problem formulations. 
% We finally provide computational experiments to ground our results (Sec.~\ref{computational_experiments}).

% \vspace{-1.5mm}
\section{RELATED WORK}
\label{sec:related-work}
% \vspace{-1mm}
\noindent Many of the current formalizations and methodologies used in modern robotics are inadequate for automating robot design as they primarily address single-domain challenges \cite{nilles_robot_2018}. Thus, specific efforts must be made to solve the problem of optimal robot co-design, which requires methods that can cross one or more levels of abstraction.

% One approach to the co-design problem uses model-free abstractions in which the design and model of the robot is not specified in advance. Instead, there is higher level goal (e.g., create a walking robot that can walk the fastest), and robots are designed (often using heuristics) to maximize performance in achieving this goal. Note that while the larger robot design itself is unspecified and does not have an explicit model, the rest of the robot and its environment are typically very well modeled to facilitate simulation- or heuristics-based approaches. 
% % Also, while the design space is largely unrestricted, the implementation space is often restricted in some manner to keep the problem tractable. 
% With this approach there are no guarantees on any produced designs being globally optimal, although model-free approaches can produce results that may be near-optimal.
% Also, while model-free approaches generally produce near-optimal results, there are no guarantees on any produced designs being globally optimal.

One approach to the co-design problem uses model-free abstractions in which the robot's design is not predetermined.
Instead, a high-level goal guides the design process to achieve the desired functionality.
Model-free approaches include evolutionary algorithms~\cite{pinskier_bioinspiration_2022, bhatia_evolution_2021}, graph heuristic approaches~\cite{zhao_robogrammar_2020}, human-assisted design~\cite{desai_computational_2017, desai_assembly-aware_2018, mehta_cogeneration_2014}, and modular approaches~\cite{jing_accomplishing_2018}.
While these methods lack a higher-level model dictating subcomponent connections, the subcomponents and environment are often well-modeled to facilitate simulation- or heuristics-based optimization.
% While these methods lack a higher-level model dictating how subcomponents should be connected, the subcomponents and environment are often well-modeled to facilitate simulation- or heuristics-based optimization.
Additionally, they do not guarantee optimal solutions.
% Additionally, there's no guarantee that generated solutions are optimal.
% Also, while the design space is largely unrestricted, the implementation space is often restricted in some manner to keep the problem tractable. 

In contrast to model-free approaches, model-based approaches define an explicit model for the entire robot design. 
These approaches are well-suited for the problem of component selection (where components must be selected from large catalogs) as the model can be used to specify component types and relationships between these components. 
While this restricts the design space, model-based design enables formal guarantees on the system. 
% While specifying component types within the model restricts the design space, formal guarantees can be placed on the system. 

The most common model-based methods are linear and quadratic programming.
% The most common methods for model-based approaches are linear and quadratic programming.
These methods leverage extensive prior research and highly optimized libraries to achieve fast component selection in \cite{magnussen_multicopter_2015} and \cite{carlone_robot_2019}.
However, designers must linearize the system, which is challenging due to the prevalence of non-linear dynamics.
% However, a designer must linearize the system, which is not trivial since non-linear system dynamics are common.

A second model-based approach is Monotone Co-Design Problems (MCDPs). MCDPs are a framework to describe multi-objective systems where monotone subsystems are connected via functionalities and resources~\cite{censi_mathematical_2015}.
They can solve multi-objective problems with non-linear, non-convex, or non-continuous systems~\cite{zardini_co-design_2021}, but require all relationships between functionalities and resources to be monotone. 
% They can solve multi-objective problems with non-linear, non-convex, or non-continuous systems~\cite{zardini_co-design_2021}, but the relationships between functionalities and resources must be monotone. 
While inspired by MCDPs, our formulation supports a broader range of constraints and focuses on scalability for large component catalogs~\cite{wilhelm_constraint_2023}.

The proposed approach and our previous work~\cite{wilhelm_constraint_2023} use constraint programming, also a model-based approach. 
% Note that we use integer programming, not \textit{linear} integer programming, which means we don't need to linearize our system.
Constraint programming works well for component selection as it is compatible with non-linear systems and can work with large component catalogs that have discrete component choices ill-suited for linearization.
It also provides a more natural representation for the problem of component selection and, as demonstrated in Sec.~\ref{computational_experiments}, scales more efficiently than binary integer linear programming.
% Constraint programming is a more general problem than linear programming, so to solve multi-objective problems techniques such as AND/OR Branch-and-Bound \cite{gent_exploiting_2009} or Russian doll search \cite{rollon_multi-objective_nodate} can be used instead of traditional linear optimization techniques.

% Our monotone subsystem decomposition technique is most similar to the hierarchical Pareto optimization approach of Singh et al. \cite{singh_hierarchical_2006}, used in their work to design sustainable industrial ecosystems. Hierarchical Pareto optimization works by first optimizing subsystems for Pareto fronts and then iteratively optimizing the larger system by updating ``interaction variables" shared between the subsystems and larger system. 
% % The larger system and subsystems share ``interaction variables'', and so after the larger system has been optimized, the subsystems are then optimized again using the new values for the interaction variables. This continues iteratively until convergence. 
% In contrast, our approach does not iteratively improve the system optimum as once a subsystem is optimized for a Pareto front, it does not need to be re-optimized.  
% Our approach requires monotonicity, but we also prove that solutions that use subsystem Pareto fronts are globally optimal. 

Our monotone subsystem decomposition technique is most similar to the hierarchical Pareto optimization approach of Singh et al., applied in~\cite{singh_hierarchical_2006} to design sustainable industrial ecosystems.
The advantage of our approach is that we do not need to iterate on solved subsystems -- once a subsystem is optimized, it is not necessary to re-optimize. 
Our subsystem decomposition requires monotonicity, but we also prove that solutions that use subsystem PFs are globally optimal.

\vspace{-0.5mm}
\section{NOTATION}
\label{sec:problem_background}
% \vspace{-0.5mm}

% \noindent In this section, we provide a brief background on single- and multi-objective constrained optimization problems (COPs). We then discuss the framework for applying constrained optimization to the problem of component selection. 
% Afterwards, we present our subsystem decomposition approach, first proving under certain conditions that the PF of a consistent subsystem can be used to determine a PF of globally optimal solutions.
% We then discuss how to address non-monotone systems and the advantages and practical considerations for using subsystems. 
% % Afterwards, we present our subsystem decomposition approach, first proving that the PF of a consistent subsystem can be used to determine a PF of globally optimal solutions, and then discussing how to handle non-monotone systems and the advantages and practical considerations for using subsystems. 
% Finally, we compare our approach to MCDPs and discuss how MCDPs impose different modeling constraints, with our monotone subsystem decomposition offering greater flexibility in certain problem formulations.

% \vspace{-1mm}
% \subsection{Multi-Objective Constrained Optimization}
% \label{subsec:multi-objective-co}
% \vspace{-1mm}

 % Optimization problems can be minimization or maximization problems. To simplify, through the rest of the text we will only refer to minimization without loss of generality.

\vspace{-0.5mm}
\subsection{Single-Objective Constrained Optimization}
\vspace{-1mm}

\noindent Single-objective COPs are a class of optimization problems that attempt to find a solution that maximizes or minimizes an objective while satisfying a set of constraints. 
In ~\cite{wilhelm_constraint_2023}, we presented a framework for using the constrained optimization representation to solve the problem of component selection, where one knows a general robot model and needs to select specific components from a large catalog.
% In our formulation, a variable $v$ is an index set for a component type, and the domain $D_v$ of that variable is the catalog entries of that component. We introduce the notion of variable properties, where $d_v\in D_v$ is a tuple with $N_{D_v}$ elements and each element in the tuple corresponds to a property that characterizes the component and is used to describe the relationships between components. 
In our formulation, we use a finite index set for variables ${V} = \{1, 2, \ldots, {N_V}\}$,
a set of finite domains (one for each variable index $v$) $\mathcal{D} = \{D_1, \ldots, D_{N_V}\}$, and 
a finite index set for constraints $C = \{1, 2, \ldots, {N_C}\}$. 
$d_v \in D_v$ are tuples of length  $N_{D_v}$, i.e.  $|d_v| = N_{D_v}$ where tuple lengths may vary across variable domains but are uniform within each domain.
Each element in the tuple $d_v$ corresponds to a component property for variable $v$, and these properties are used to characterize the component and describe relationships between other components.
We index the variable properties $d_v$ using an index set ${P}_{v}=\{1,\ldots,N_{D_v}\}$, defined for each $v \in V$, so that the \textit{p}-th property of a variable \textit{v} can be written $D_{v,p}$.
% Similarly, let  $C = \{1, 2, \ldots, {N_C}\}$ be a finite index set for constraints that specify valid combinations of values. 

% For discrete optimization, a single-objective COP has a finite index set for variables ${V} = \{1, 2, \ldots, {N_V}\}$, 
% a set of finite domains (one for each variable index $v$) $\mathcal{D} = \{D_1, \ldots, D_{N_V}\}$. 
% The elements of each $D_v$ are tuples of length  $N_{D_v}$, i.e.  $|d_v| = N_{D_v}$ for elements $d_v \in D_v$. 
% The tuples in each variable domain can have different lengths, but all tuples in a domain are the same length.
% % a set of finite domains (one for each variable $V_i$) $D_i = \{d_{i,1}, d_{i,2}, \ldots, d_{i,m}\}$, 
% We denote a specific assignment of all variables to specific domain values as $\mathbf{v}=(d_1, d_2, ..., d_{N_V})$ and the set of all possible assignments as  \textbf{V}.

Constraints are mathematical expressions that describe relationships between the properties of components. 
These constraints can not only specify valid component combinations but also describe the performance specifications of the robot.
Inequality constraints take the general form of $L_c(D_{v,p},D_{v',p'},...)\leq R_c(D_{v,p},D_{v',p'},...)$  where $L_c:\mathbb{R}^{\sum_{v}|P_v|}\rightarrow\mathbb{R}$ and $R_c:\mathbb{R}^{\sum_{v}|P_v|}\rightarrow\mathbb{R}$ are real-valued functions that operate on the variable properties and describe constraint $c \in C$. 

The goal is to find an assignment $\mathbf{v}$ that maximizes or minimizes an objective function $f_i(\mathbf{v})$ subject to the constraints.
Objective functions can also be defined on variable properties $f_i(\mathbf{v}) \equiv f_i(D_{v,p},D_{v',p'},...):\mathbb{R}^{\sum_{v}|P_v|}\rightarrow\mathbb{R}$, and we denote $f_i^{\downarrow}$ and $f_i^{\uparrow}$ according to whether these objective functions should be minimized or maximized, respectively.
% The goal is to find an assignment $\mathbf{v}$ that maximizes or minimizes a real-valued objective function $f_i(\mathbf{v}) \equiv f_i(D_{v,p},D_{v',p'},...)$ subject to the constraints.
Note we use a lowercase $v$ for indexing and a bold $\mathbf{v}$  for assignments.
% The goal is to find a solution $d_V=(d_1, d_2, ..., d_{N_V})$ that maximizes or minimizes an objective function $f: D_1 \times \cdots \times D_n \to \mathbb{R}$ subject to the constraints.
% An objective function $f(v_i,\ldots)$ maps a combination of variables to a real value. 
Solving a COP can be formulated as a backtracking tree-search, see our previous work for more details~\cite{wilhelm_constraint_2023}.

To aid development of the constraint-based model, it is often easier to use descriptive names, and we associate finite strings with elements of the different index sets. Let the mappings $h_{V}:{V} \rightarrow \Sigma^*$, $h_{{{P}_v}}:{P}_v \rightarrow \Sigma^*$, and  $h_{{C}}:C \rightarrow \Sigma^*$ associate unique names (finite strings over a finite alphabet $\Sigma$) to variables, their properties, and constraints, improving readability when writing equations.
For example, in our quadcopter problem, a battery variable represents different battery choices and has properties such as current and voltage.
Using these mappings, we write $D_{v,p}$ as $B_{\textrm{voltage}}$, where $D_{v,p} = D_{h^{-1}_V(B), h^{-1}_{P_v}(\textrm{voltage})}$ and $h^{-1}_V$ and $h^{-1}_{P_v}$ are the inverse mappings of $ h_V(v) = B $ and $ h_{P_v}(p) = \textrm{voltage} $.
% This follows from \( h_V(v) = B \) and \( h_{P_v}(p) = \textrm{voltage} \), making constraints more readable.
% With these mappings, we can represent the variable property as $B_\textrm{voltage}$, where $D_{v,p}=D_{h^{-1}_{V}(B), h^{-1}_{P_v}(\textrm{voltage})}$is represented as $B_\textrm{voltage}$, where $h_{V}(v)=B$ and $h_{P_v}(p)=\textrm{voltage}$. 
% With these mappings, $D_{v,p}$ is represented as $B_\textrm{voltage}$, where $h_{V}(v)=B$ and $h_{P_v}(p)=\textrm{voltage}$. 
This shorthand allows constraints to be conveniently expressed as:
% We would write out a \textit{battery_power_constraint} 
% For example, in our quadcopter problem we have a camera variable $V_\textrm{camera}$ that represents the choices of different cameras, and this variable has properties such as $mass$ and $frame\_rate$.
\vspace{-1.75mm}
\[4 * (M_{\textrm{voltage}} * M_{\textrm{current}})\vspace{-1.25mm} \leq B_{\textrm{voltage}} * B_{\textrm{current}}\]
% \[4 * (p_{\textrm{M,voltage}} * p_{\textrm{M,current}})\vspace{-1.25mm} \leq p_{\textrm{B,voltage}} * p_{\textrm{B,current}}\]
where the terms of the left and right hand side refer to the associated voltage and current properties of the motor (M) and battery (B) variables, respectively.

\vspace{-1mm}
\subsection{Multi-Objective Constrained Optimization}
\vspace{-0.5mm}

Multi-objective optimization aims to simultaneously optimize several objective functions to obtain an efficient front of solutions (i.e., a PF). 
% As opposed to single-objective optimization, multi-objective optimization simultaneously aims to optimize several objective functions to obtain an efficient front of solutions (equivalently, a PF). 
A vector of objective functions ${F}(\mathbf{v})=(f_1(\mathbf{v}),...,f_{N_F}(\mathbf{v}))$ represents the quantities to be optimized, with a corresponding index set $I=\{1,\ldots,N_F\}$. 
% As opposed to single-objective optimization, the goal of multi-objective optimization is to simultaneously optimize more than one objective function to find an efficient front of solutions (equivalently, a Pareto front). A vector of objective functions $F(v)=(f_1(v),...,f_{N_f}(v))$ represents the quantities to be optimized. 
This induces a partial order on the search space, where (assuming minimization $\forall f_i \in F$) an assignment $\mathbf{a}$ ``dominates" another assignment $\mathbf{b}$, i.e., $\mathbf{a} \preceq \mathbf{b}$, if $f_i(\mathbf{a}) \leq f_i(\mathbf{b}), \forall i \in I$ and at least one inequality is strict.
%\wedge \exists j \in {1, \ldots, o} : f_j(y) < f_j(x)$ % math version of what I write out
The optimal PF $O_V$ is the set of non-dominated solutions $\{\textbf{v} \in \textbf{V}\ |\  \nexists \textbf{v}' \in \textbf{V},  \textbf{v}' \preceq \textbf{v} \}$.
% The Pareto front $P_V$ is the set of solutions $\{v \in V : v' \preceq v, v' \notin P_V\}$.
%The Pareto front is an anti-chain of optimal solutions to the problem where no point on the Pareto front dominates another point.  A set of points closed under non-domination is called a frontier.

To solve a multi-objective optimization problem, a solver maintains an upper bound frontier and computes a lower bound frontier, backtracking if the upper bound dominates the lower. 
More advanced methods (e.g. AND/OR branch-and-bound \cite{gent_exploiting_2009} or Russian doll search \cite{rollon_multi-objective_2007}) efficiently explore the search space. 
% More advanced methods (e.g. AND/OR branch-and-bound \cite{gent_exploiting_2009} or Russian doll search \cite{rollon_multi-objective_2007}) efficiently explore the search space which we do not go into detail here for space reasons. 
However, widely available industrial solvers such as IBM CPLEX~\cite{noauthor_ibm_nodate} and Gurobi~\cite{noauthor_gurobi_nodate} do not perform true multi-objective optimization (instead using lexicographical or weighted objectives).  

In this paper, we use a different technique to determine the PF of solutions, choosing to take advantage of these highly optimized libraries (in particular IBM CPLEX). We use $k$th-objective $\epsilon$-constraint scalarization~\cite{chiandussi_comparison_2012}, where one objective $f_k$ is minimized while the rest are constrained by upper bounds $f_i(x)<\epsilon, \forall i \in I, i \neq k$. 
In our case, we choose upper bounds of previously found solutions and perform a lexicographical single-objective solve, adding a constraint to ensure new solutions improve at least one objective (i.e. a new solution must not be dominated by the current PF). 
% In our case, we choose upper bounds of previous found solutions and perform a lexicographical single-objective solve, adding a constraint that enforces that at least one of the objectives of a newly found solution must be better than previous solutions (i.e. a new solution must not be dominated by the current PF). 
We iteratively solve for new solutions, adding new constraints until no new solutions are found. 
The resulting set of optimal solutions forms a PF. 
%Note that a $k$th-objective $\epsilon$-constraint scalarization approach is better than minimizing the weighted sum of the objectives, as the latter cannot find optimal points lying in non-convex sections of the frontier.

% In this paper, we use a different technique to determine the PF of solutions, choosing to take advantage of these highly optimized libraries (in particular IBM CPLEX). We use $k$th-objective $\epsilon$-constraint scalarization~\cite{chiandussi_comparison_2012}, where one objective $f_k$ is minimized while the rest are constrained by upper bounds $f_i(\textbf{v})<\epsilon, \forall i \in I, i \neq k$. In our case, we choose upper bounds of previous found solutions and perform a lexicographical single-objective solve, adding a constraint that enforces that at least one of the objectives of a newly found solution must be better than previous solutions (i.e. a new solution must not be dominated by the current PF). We iteratively solve for new solutions, adding new constraints until no new solutions are found. 

\section{PROBLEM FORMULATION}
\label{sec:problem_formulation}

 % \vspace{-4mm}
% \subsection{Constrained Optimization for Component Selection}
 % \vspace{-3mm}
\noindent Using the notation from Sec.~\ref{sec:problem_background}, we now describe the foundational framework for describing monotone subsystem decomposition. 
% \noindent Using the notation from Sec.~\ref{sec:problem_background}, we now describe how component relationships can be represented using a bipartite multi-graph and the foundational framework for describing monotone subsystem decomposition. 
% \noindent Using the notation from Sec.~\ref{sec:problem_background}, we now describe how component relationships are represented using a bipartite multi-graph and the foundational framework for describing monotone subsystem decomposition. 

\begin{figure}[t]
      \vspace{2mm}
      \centering
      
      \includegraphics[scale=0.18]{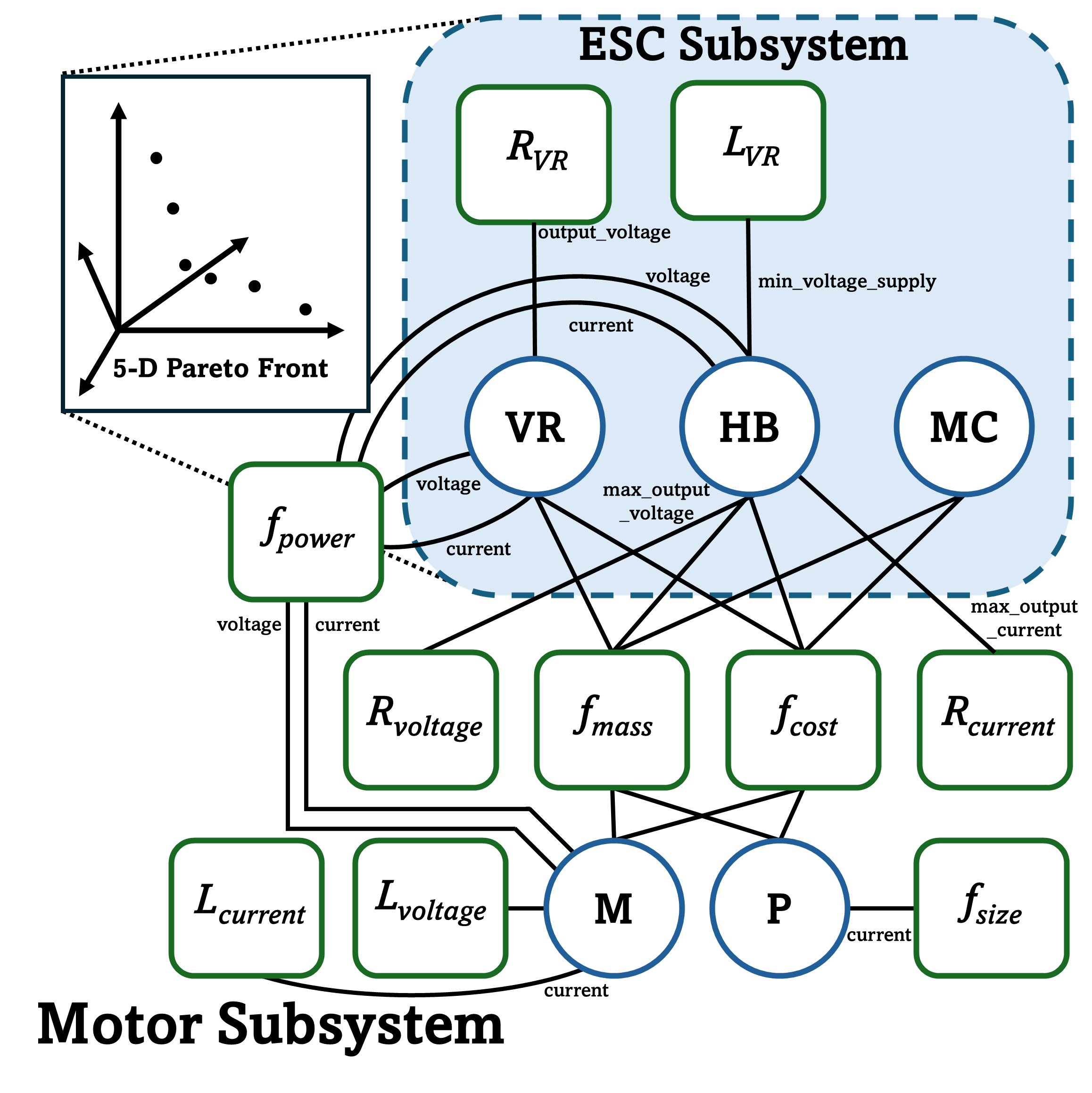}
                        \vspace{-5mm}
      \caption[Caption for LOF]{\small A constraint graph for the Motor and ESC subsystems in Sec.~\ref{computational_experiments}. The voltage regulator (VR), half bridge (HB), microcontroller (MC), motor (M), and propeller (P) components are represented as circles, and constraint and objective functions as squares. 
      An edge connects a variable to a function if that variable has a property in the domain of the function; select edges and their properties have been labeled. 
      A subsystem (blue dashed line) is a subset of component nodes. 
      % A subsystem (blue dashed line) is a subset of component nodes and the functions that are only used within that subset. 
      If fully consistent, the subsystem's PF can be used to optimize the larger system. 
      % For the ESC subsystem, the mass, cost, power, max output voltage, and max current must be optimized.
      }

      \label{constraintGraphFig}
    \vspace{-6.5mm}
  \end{figure}

To keep track of constraint and objective function dependencies on variables, we use a bipartite multi-graph. 
One set of nodes represents variables and the other set represents functions of the properties, see Fig.~\ref{constraintGraphFig}. An edge between a function and a variable indicates that the function depends on a property of that variable. 
Parallel edges indicate a function has multiple properties of the same variable in its domain.

When functions are consistently increasing or decreasing, subsystems can be optimized independently and their results used to optimize the larger system.
A variable property is \textit{monotone} if increasing its value never decreases the function value, and \textit{antitone} if increasing its value never increases the function value.
A variable property is \textit{consistent} if it is either a \textit{consistent maximization property} $D^\uparrow_{v,p}$ or a \textit{consistent minimization property} $D^\downarrow_{v,p}$.
A variable property is a \textit{consistent maximization} property $D^\uparrow_{v,p}$ iff it appears exclusively in inequality constraints where it is antitone for $L_c$ and $f_i^{\downarrow}$ and monotone for $R_c$ and $f_i^{\uparrow}$. 
Likewise, a variable property is a \textit{consistent minimization }property $D^\downarrow_{v,p}$ iff it appears exclusively in inequality constraints where it is monotone for $L_c$ and $f_i^{\downarrow}$ and antitone for $R_c$ and $f_i^{\uparrow}$.
A constraint function $L_c$, $R_c$, or objective function $f_i$ is \textit{fully consistent} if all variable properties in its domain are consistent.

We can determine a subsystem ${X}$ within the constraint graph by selecting a subset of variables from the larger system ${Z}={X}\cup {Y}$, where ${X} \subset {Z} \subseteq {V}$ and ${Y} = {Z}\setminus {X}$. 
% An implementation of $X$ is written $x_i = \{v_{q,i} \}$
We denote an implementation $\mathbf{z}$ of subsystem ${Z}$ that uses implementations $\mathbf{x}$ and $\mathbf{y}$ from ${X},{Y}$  as $\mathbf{z}=\mathbf{x}\cup \mathbf{y}$.
A system ${X}$ shares a constraint or objective function with a system ${Y}$ if a variable property from both ${X}$ and ${Y}$ is included in the domain of that constraint or function. 
We call a subsystem \textit{fully consistent} if all constraint and objective functions that are shared between ${X}$ and ${Y}$ are fully consistent. 

\vspace{-1mm}
\section{THEORETICAL RESULTS}
\label{sec:theoretical_results}
\vspace{-1mm}
% \subsection{Subsystem Optimization for Constraint Programming}
  
\noindent In this section, we prove that optimizing a consistent subsystem and using its PF to optimize the larger system still yields globally optimal results, the basis for our monotone system decomposition technique.
This accelerates computation as we can precompute subsystem solutions and leverage this reduced set to efficiently solve the larger design problem, drastically reducing the search space. 
% This accelerates computation by drastically reducing the search space of the larger design problem. 
% By precomputing subsystem solutions, we can leverage the reduced set of implementations from the subsystem's PF to determine globally optimal solutions to the larger problem.
% We can precompute solutions of the subsystem and then use the reduced set of implementations from the subsystem's PF to determine globally optimal solutions to the larger problem.

% \subsection{Proof of Global Optimality in Optimizing Consistent Subsystems}

% \textbf{Theorem:} \textbf{(Global Optimality in Optimizing Consistent Subsystems)}
\textit{\textbf{Theorem:}}
\textit{Consider the optimization of a system ${Z}={X}\cup {Y}$ where  ${Z} \subseteq {V}$ and ${X}$ is a fully consistent subsystem within ${Z}$. Then all assignments of ${Z}$ on the Pareto front $O_Z$ can use assignments from the Pareto front $O_X$ of the subsystem ${X}$, where $O_X$ is optimized for variable properties used in shared functions between ${X}$ and ${V}\setminus {X}$.}

Subsystem consistency ensures that variable properties are optimized consistently, minimizing $f_i^{\downarrow}$ and $L_c$ and maximizing $f_i^{\uparrow}$ and $R_c$. 
% Subsystem consistency ensures that when we optimize the variable properties, they are optimized in a consistent manner such that $f_i^{\downarrow}$ and $L_c$ functions are minimized and $f_i^{\uparrow}$ and $R_c$ functions are maximized. 
This guarantees that, for any $\mathbf{x'} \notin O_X$, there is an assignment $\mathbf{x} \in O_X$ that will have better objective values and will satisfy the same inequality constraints in $Z$.
% With this property, we can precompute solutions of the subsystem and then use the reduced set of implementations from the subsystem's PF to determine globally optimal solutions to the larger problem.

\textit{\textbf{Proof}}: As ${X}$ is a fully consistent subsystem, its shared properties are either maximization or minimization properties.
% As ${X}$ is fully consistent, its shared properties are either maximization or minimization properties 

Consider the maximization property $D^\uparrow_{x,p} \in D_x$ in $X$. Let $\mathbf{x'}$ and $\mathbf{x}$ be assignments such that $\mathbf{x'} \notin O_X$,  ${\mathbf{x}} \in O_X$. 
Thus, $d^\uparrow_{\mathbf{x'},p} \leq d^\uparrow_{\mathbf{{x}},p}$ for $d^\uparrow_{\mathbf{x'},p}, d^\uparrow_{\mathbf{{x}},p} \in D^\uparrow_{x,p}$ as $O_X$ was maximized for $D^\uparrow_{x,p}$. 
% Thus, $d^\uparrow_{\mathbf{x'},p} \leq d^\uparrow_{\mathbf{{x}},p}$ as $O_X$ was maximized for $D^\uparrow_{x,p}$. 
Additionally, let $f^\uparrow_i(D_{{x},p}, D_{y,p'}, \ldots)$ and $f_i^\downarrow(D_{{x},p}, D_{y,p'}, \ldots)$ represent objective functions in the system ${Z}$, and let $R_c(D_{x,p},D_{y,p'},...)$ and $L_c(D_{x,p},D_{y,p'},...)$ represent shared constraint functions in ${Z}$.

We first prove optimality, in that an assignment $\mathbf{z}=\mathbf{x} \cup \mathbf{y}, \mathbf{y} \in \mathbf{Y}$  will dominate an assignment $\mathbf{z'}=\mathbf{x'} \cup \mathbf{y}$. We then prove feasibility, in that a solution $\mathbf{z}$ will satisfy all constraints met by $\mathbf{{z'}}$.

\textbf{Optimality:} Since the objective functions $f_i^{\uparrow}$ and $f_i^{\downarrow}$ are fully consistent and $D^\uparrow_{x,p}$ is a maximization property, it follows $f^\uparrow_i(d^\uparrow_{\mathbf{x'},p}, d_{\mathbf{y},p'}, \ldots) \leq f^\uparrow_i(d^\uparrow_{\mathbf{x},p}, d_{\mathbf{y},p'}, \ldots)$ and $f^\downarrow_i(d^\uparrow_{\mathbf{x},p}, d_{\mathbf{y},p'}, \ldots) \leq f^\downarrow_i(d^\uparrow_{\mathbf{x'},p}, d_{\mathbf{y},p'}, \ldots)$.

\textbf{Feasibility:} Since the constraints are structured $L_c(D_{x,p},D_{y,p'},...)\leq R_c(D_{x,p},D_{y,p'},...)$, then $\mathbf{x}$ will not violate any constraints that $\mathbf{x'}$ does not already violate since $L_c(d^\uparrow_{\mathbf{x},p},d_{\mathbf{y},p'},...)\leq L_c(d^\uparrow_{\mathbf{x'},p},d_{\mathbf{y},p'},...)$ and $R_c(d^\uparrow_{\mathbf{x'},p},d_{\mathbf{y},p'},...) \leq R_c(d^\uparrow_{\mathbf{x},p},d_{\mathbf{y},p'},...)$ as $R_c$ and $L_c$ are fully consistent.

Thus, for a shared maximization property $D^\uparrow_{x,p}$, we have proven for any $\mathbf{z'}=\mathbf{x'}\cup \mathbf{y}, \mathbf{x'} \notin {O_X}, \mathbf{y} \in {\mathbf{Y}}$, there is an assignment ${\mathbf{z}}=\mathbf{x} \cup \mathbf{y}, \mathbf{x} \in {O_X}$ such that ${\mathbf{z}} \preceq \mathbf{{z'}}$.
The above holds for all $D^\uparrow_{x,p}$ shared between $X$ and $Y$. For minimization properties $D^\downarrow_{x,p}$, the proof is complementary but omitted due to space limitations.
Since ${\mathbf{z}} \preceq \mathbf{{z'}} \forall \mathbf{x} \in O_X, \mathbf{x'} \notin O_X$, and by definition $O_Z=\{\mathbf{z} \in \mathbf{Z} | \nexists \mathbf{z'} \in \mathbf{Z},  \mathbf{z'} \preceq \mathbf{z} \}$, then ${O}_{Z}$ will use assignments from ${O_X}$. \hfill $\blacksquare$

\vspace{-1mm}
\section{DISCUSSION}
\label{sec:discussion}
\vspace{-1mm}
\noindent In this section, we highlight advantages of monotone subsystem decomposition, address optimization of non-monotone systems, provide practical considerations, and compare our approach to MCDPs.
% and discussing how MCDPs impose different modeling constraints, with our monotone subsystem decomposition offering greater flexibility in certain problem formulations

\subsection{Advantages of Using Subsystems for Optimization}
\noindent Using subsystems for optimization provides several advantages. First, subsystems are an intuitive way of describing the larger problem, and this framework is already commonly used by engineers to build complex systems.

Second, optimizing subsystems can provide a designer with internal information that can be helpful for system design. For example, in examining the PF of a sensing/computing subsystem, a designer may realize that a larger design problem is failing to meet design specifications since the subsystem does not have enough computing power. 

Third, subsystems can be reused within the same or across different design problems, saving on computation. 
If the larger design problem changes or the subsystem is applied to a different design problem, the subsystem's PF can be reused as long as its minimization and maximization properties remain unchanged.
Furthermore, subsystems replicated in a larger design system (e.g. quadcopter designs used in a quadcopter delivery problem, see Sec.~\ref{quadcopter_fleet_results_sec}) can be optimized once and the resulting PF used as an ``aggregate" component in the larger system, avoiding redundant computations. 
% Furthermore, subsystems that are replicated in a larger design system (e.g. quadcopter designs used for a quadcopter delivery problem, see Results Sec.~\ref{quadcopter_fleet_results_sec}) can be optimized once, and the resulting PF used as an ``aggregate" component in the larger design system, rather than optimizing the identical problem at the subsystem level several times over within the larger problem. 
% Subsystems can also be reused in completely separate design problems, as long as the minimization and maximization properties are the same between problems.

\subsection{Optimizing Non-Monotone Systems}
\label{optimizing_non_monotone}

% Systems that have constraint and objective functions that are not fully consistent (e.g. non-monotone relationships or equality constraints) can still be optimized using the above methodology, after some considerations. 
\noindent Systems containing constraint or objective functions that are not fully consistent (e.g., non-monotone relationships or equality constraints) can still be optimized with this method after additional considerations. 

First, we note that functions that are not fully consistent only violate the assumptions of the above proof if they are shared between the subsystem and the larger system. 
Inconsistent functions internal to the subsystem or exclusive to the larger system can still be optimized using standard CP methods without requiring consistency.
% If they are internal to the subsystem, then the subsystem can be optimized using standard CP as this does not require consistency. 
% Similarly, this is true for the larger system if these functions are not shared with the subsystem.

% Otherwise, if these inconsistent functions involve properties shared between the subsystem and the larger system, and no other information on relationships, constraints, or properties is available, then every possible value of that property must be evaluated. 
Otherwise, if these inconsistent functions are shared with the subsystem and assuming no other information on relationships, constraints, or properties, then every possible value of the inconsistent property must be evaluated. 
For example, if logic voltages appear in equality constraints, the system must be optimized for each logic voltage in the catalog.
If the property is drawn from a relatively small finite set, the resulting PF may remain manageable. 
However, if the property spans a large finite set, possibly all subsystem implementations could be represented on the PF. 
Thus, selecting the proper subsystem is critical for tractability.

% Otherwise, if these inconsistent functions are shared with the subsystem, assuming no other information on relationships, constraints, or properties, then all possible values of that property must be evaluated in any constraints or relationships it is involved in.
% For example, if working with logic voltages that are used with equality constraints, then the problem must be optimized to include each possible logic voltage in the system. 
% If the property is discrete, then the PF may not grow too large in size. 
% However, for a continuous property, possibly all implementations of the subsystem will be represented on the PF. 
% Thus, selecting the proper subsystem to optimize is imperative to keeping runtime tractable and search spaces small. 

\subsection{Practical Considerations for Using Subsystems}

% Selecting the proper subsystem is imperative to keeping runtime tractable.
\noindent We provide guidelines for selecting effective subsystems, but in general this task depends on the details of a specific problem. First, an ideal subsystem has as few external properties as possible. 
Fewer external properties results in fewer optimization objectives, which is important for tractability as the solution space for many-objective problems scales rapidly. 
Second, it is easiest and most insightful for a designer to choose intuitive subsystems (e.g., the powertrain of a mobile robot) as these are more likely to be used in other systems (e.g., most robot designs would like to maximize the efficiency of a powertrain while minimizing cost). Finally, we note that even a single component can be made into a subsystem, which can be extremely beneficial as that component's catalog can be optimized for a PF before other computation, potentially reducing the search space of the larger design problem.

% What happens if a designer misclassifies a property, e.g. labels a resource as a or vice versa? In the first case, the run time will be longer and the solutions space much larger, but any final results of the larger design problem will be unaffected. In the second case, where an incomparable is labeled a resource, the search time would be faster, but using the resulting Pareto front in the larger design problem would no longer necessarily yield optimal results. However, these sub-optimal results still may be very close to optimal results depending on the system.

% \vspace{-2mm}
\subsection{Comparison of Our Approach to MCDPs}
% \vspace{-2mm}
\noindent Our CP approach with monotone subsystem decomposition is similar to MCDPs in that they both use monotonicity and can solve multi-objective optimization problems; however, they differ in a few aspects. 
First, MCDPs can easily express continuous and discrete domains while our approach is limited to discrete domains but is better suited for the discrete problem of component selection~\cite{wilhelm_constraint_2023}. 
% Second, our monotone subsystem decomposition technique places fewer restrictions on the model. 
Second, our monotone subsystem decomposition imposes fewer models restrictions. 
MCDPs require all relationships between variables to be monotone.
Our approach allows both monotone and antitone relationships without requiring monotonicity throughout the entire system.
% Our approach only requires monotonicity that can be monotone or antitone, and monotonicity is not necessary throughout the entire system. 
In our approach's framework, a discrete domain MCDP would be a consistent system with only maximization properties.
Finally, CP offers the advantage of interfacing with other traditional CP problems, such as packing or scheduling (as demonstrated in the quadcopter fleet design problem in Sec.~\ref{quadcopter_fleet_results_sec}).

% \vspace{-2mm}
\section{Computational Experiments}
% \vspace{-1mm}
\label{computational_experiments}
% \vspace{-2.5mm}

\noindent In this section, we first design a quadcopter system and compare our CP approach to a linear programming approach, the current state-of-the-art in terms of computational speed. We then expand upon this quadcopter design problem and solve a multi-objective, task-oriented scheduling problem to design a fleet of quadcopters to complete package deliveries as cheaply and as quick as possible. 
% We solve a large scale scheduling problem with X drones and Y packages, providing optimal package delivery schedules and globally optimal drone designs down to the component level.

In practice, due to the scale of our catalogs, there can exist several alternative solutions at a single point on the PF. We provide only one solution per point under the assumption that if a specific point interests a designer, then all solutions can be returned for that point with further processing. 

% All provided run times in this section were collected over 10 trials on a standard office desktop computer (2.10 GHz CPU with 12 cores, 16 GB RAM).
All provided run times in this section were collected on a standard office desktop computer (4 GHz CPU with 12 cores, 32 GB RAM).

\vspace{-1.5mm}
\subsection{Quadcopter System Design Problem}
\vspace{-1mm}
\label{quadcopter_results_sec}
\noindent We compare our CP approach to the LP method of Carlone et al.~\cite{carlone_robot_2019}. We use the same quadcopter design problem, a single-objective problem to design a quadcopter of maximum velocity within a fixed budget of \$1k. The quadcopter is composed of five types of components: motor, computer, camera, battery, and frame (see~\cite{carlone_robot_2019} for full model and constraints). 
% While we encourage readers to refer to \cite{carlone_robot_2019} for the full model and constraints, we highlight here that there are non-monotone constraints within the model. 
We highlight that there are non-monotone constraints within the model. 
For example, velocity is a non-monotone quantity: the goal is to maximize velocity as that is the primary design objective, but within the constraints this quantity needs to be minimized since the camera’s minimum frame rate and the computer’s minimum back-end throughput are directly proportional to the velocity. 

Carlone et al. use catalogs that have 19, 4, 4, 14, and 6 items for the motor, computer, camera, battery, and frame, respectively. 
To better assess scalability of each approach, we generate synthetic catalogs of 100,000 components per type based on the mean and standard deviation of the real-world components from the original catalogs.
% To better capture the scaling of each approach, we generate synthetic catalogs of 100,000 components per type based on the mean and standard deviation of the real-world components from the original catalogs.
This yields problem sizes of up to 10\textsuperscript{25} component combinations.

\begin{figure}[t]
\vspace{3mm}
      \centering
      
      \includegraphics[scale=0.5]{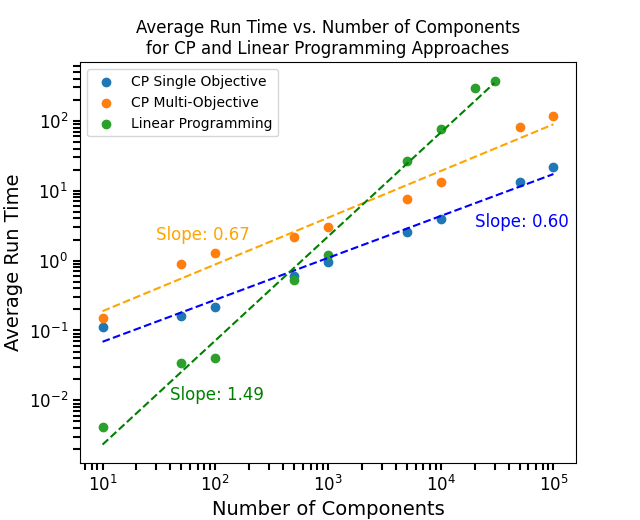}
      \caption[Caption for LOF]{\small Comparison of our CP approach versus the linear programming approach. At larger catalog sizes, the CP produces solutions faster than LP, and the multi-objective CP even produces an entire PF faster than the time it takes LP to determine one single-objective solution. Above 30,000 components, the LP application crashes.}

      \label{CPversusLPFigure}
    \vspace{-5mm}
  \end{figure}

Figure~\ref{CPversusLPFigure} compares the scaling performance of our CP approach against the LP approach. 
While the original quadcopter design problem is single-objective, our method extends to multi-objective optimization so we also report timing results for solving a multi-objective variant that minimizes system mass alongside maximizing velocity.
% Since the original quadcopter design problem is single-objective but our approach can solve multi-objective problems, we also provide timing data solving a multi-objective problem of the same quadcopter design but minimizing system mass in addition to maximizing maximum velocity. 
As shown, our CP approach produces solutions faster than LP for all but the smallest of catalogs (in which solve times are less than a second).
For catalogs exceeding 200 components, our CP approach even solves the multi-objective design problem and determines a PF of optimal solutions faster than the LP approach can solve the single-objective case. 
Our approach has a log-log slope $<1$, indicating its scaling is sub-linear.

\subsection{Quadcopter Fleet Design Problem}
\label{quadcopter_fleet_results_sec}

\noindent We then applied our work to a larger and more complex case study solving the multi-objective, task-oriented scheduling problem of building a fleet of quadcopters and assigning them to deliver packages, all while optimizing for the delivery time and cost to build the fleet. 
We also solve this problem with a single-objective (only minimizing cost) to showcase the speed and scalability of our approach.

\begin{figure}[t]
\vspace{2mm}
      \centering
      
      \includegraphics[scale=0.30]{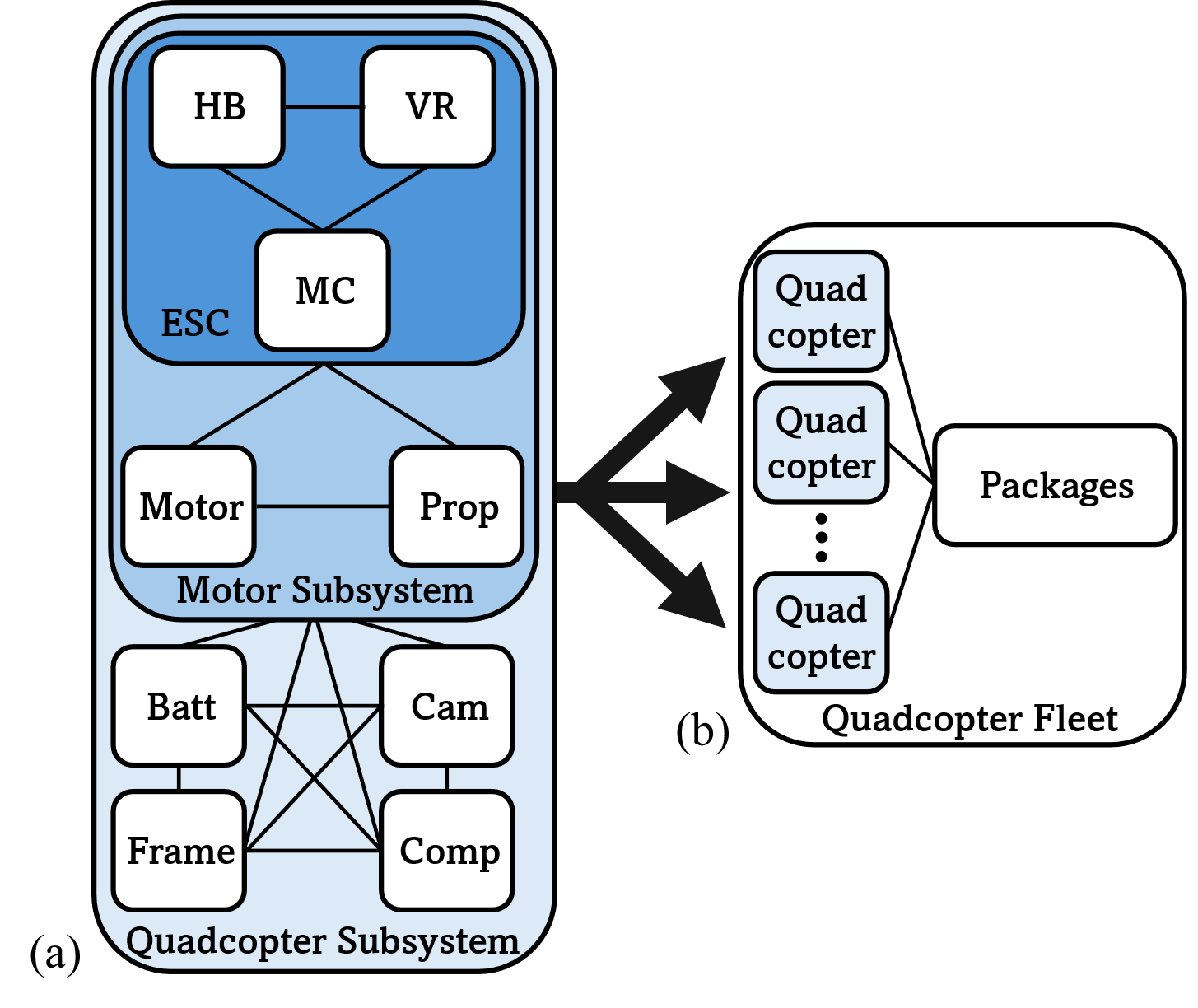}
      \caption[Caption for LOF]{\small System diagrams for the quadcopter (a) and quadcopter fleet (b) design problems. The quadcopter contains a motor subsystem, which itself has an ESC subsystem. Several instantiations of the quadcopter serve as subsystems in the quadcopter fleet design problem, where each quadcopter is assigned packages for delivery.}

      \label{fleetSystemFig}
    \vspace{-5mm}
  \end{figure}

This larger fleet design problem is broken up into several smaller design problems, see Figure~\ref{fleetSystemFig}. The highest level of abstraction entails the quadcopter and package assignment. Quadcopters are assigned to deliver packages. Packages all go to unique locations, and after delivery a quadcopter must return to the depot to pick up another package for delivery. A maximum number of unique quadcopter designs can be used; however, if more than one quadcopter design is chosen, an extra financial cost is incurred for each additional design.

The second level of abstraction is the quadcopter subsystem, nearly identical from Sec~\ref{quadcopter_results_sec}, with two adjustments. 
% However, we make two adjustments. 
First, the quadcopter constraints formulated by~\cite{carlone_robot_2019} assume a racing quadcopter; since we are performing package delivery, we modify the camera and computer constraints to better reflect that these quadcopters will be built for landing at low speeds, not cornering at high speeds.
% First, the quadcopter constraints formulated by~\cite{carlone_robot_2019} assume the quadcopter is being built for racing; since we are performing package delivery, we adjust the camera and computer constraints to better reflect that these quadcopters will be built for landing at low speeds, not cornering at high speeds.
Second, the motors of the original quadcopter subsystem are replaced with a new motor subsystem, the third level of abstraction. 
Carlone et al. assumed the motor, propeller, and electronic speed controller (ESC) were all integrated into a single ``motor" component. 
% Here, we break this out as a subsystem, which is composed of a motor, propeller, and ESC. 
Here, we break this out as a motor subsystem composed of these three components.
The ESC itself is also a subsystem and the fourth level of abstraction. It is composed of an H-bridge, microcontroller, and voltage regulator. 
Each of these subsystems has several constraints that ensure component compatibility and adherence to target specifications.

First, to demonstrate the computational advantage of using subsystems, we recorded solve times for the quadcopter subsystem with and without subsystem decomposition. 
The quadcopter subsystem can be broken down into 11 subsystems: the ESC and motor subsystems, and one subsystem for each of the 9 components in the ESC, motor, and quadcopter subsystems. 
We used catalogs of 1,000, 250, and 1,000 components for each component in the ESC, motor, and quadcopter subsystems, respectively. 
With monotone subsystem decomposition, it takes 714.3 seconds to solve for the quadcopter subsystem PF, a 93\% speed-up compared to the 1375.1 seconds to solve without subsystem decomposition.

Next, we separately optimized each subsystem in Fig.~\ref{fleetSystemFig}  and then used these results to optimize the larger design problem. Table~\ref{tab:system_solve_time} shows the number of items in the catalogs for each component type and the solve times for each subsystem. It takes 167.1 seconds to optimize for all of the subsystems, and these results can be subsequently reused in both the single- and multi-objective design problems to solve them in 1.3 and 352.9 seconds, respectively.
% It takes 62 and 516 seconds to solve the single and multi-objective design problems, respectively. Note that the subsystem optimization results can be reused for both of these problems.

\begin{table}[h]
    % \vspace{2.5mm}
    \centering
    \begin{tabular}{|c|l|c|c|c|}
        \hline 
        \textbf{System}  && \textbf{\makecell{Subsystem/\\Component}}& \textbf{\makecell{Number of \\ Components}}          & \textbf{\makecell{System \\Solve Time}}           \\ \hhline{|=|=|=|=|=|}
        Electronic  &\parbox[t]{2mm}{\multirow{9}{*}{\rotatebox[origin=c]{90}{Robot Design}}}& Voltage Reg.  & 1000                               & 43.4 sec\\ \cline{3-4} Speed  && H Bridge           & 1000 & 309 solns
        \\ \cline{3-4}  Controller && $\mu$-controller    & 1000                               &                                       
        \\ \cline{1-1}\cline{3-5}
        Motor                     && ESC                & 309 & 102.7
 sec\\ \cline{3-4} && Propeller          & 250&  179 solns\\ \cline{3-4} && Motor               & 250& 
        \\ \cline{1-1}\cline{3-5}
        Quadcopter                && Motor Subsystem              & 179& 21.0 sec\\ \cline{3-4} && Battery            & 500& 32 solns\\ \cline{3-4} && Frame              & 500& 
        \\ \cline{3-4} && Computer           & 500&                                       \\ \cline{3-4} && Camera             & 500&                                       
        \\ \hline
        Quadcopter  &\parbox[t]{2mm}{\multirow{6}{*}{\rotatebox[origin=c]{90}{Planning}}}& Quadcopter& 25&1.3 sec\\
 \cline{3-4}Fleet && Packages& 125&1 soln\\
 \cline{3-4}Single-Obj. && Designs& 10 of 32&
 \\ \cline{1-1}\cline{3-5}
        Quadcopter  && Quadcopter& 4&352.9 sec\\
 \cline{3-4}Fleet && Packages& 12&6 solns\\
 \cline{3-4} Multi-Obj. && Designs& 3 of 32&\\ \hline
    \end{tabular}
    \caption{System Components and Solve Times}
    \label{tab:system_solve_time}
    \vspace{-7mm}
\end{table}

\vspace{-2mm}
\section{CONCLUSION}
\label{conclusion}
\vspace{-1mm}
\noindent In this paper, we introduce a novel CP-based technique to optimize subsystems and leverage these PFs to efficiently solve larger multi-objective design problems, ensuring global optimality and providing useful design insights about internal subsystem design choices.
% This approach keeps the otherwise large design problem tractable and provides a designer with useful insight about internal subsystem design choices.
% While the decomposition relies on monotonicity of subsystems, CP can solve more general problems and we can exploit monotonicity for computational gain when it is present.
Although this decomposition exploits monotonicity for computational gains, CP can still handle more general problems when monotonicity is absent.
We demonstrate that our approach outperforms the previous state-of-the-art by an order of magnitude, and applied it to a larger, task-oriented scheduling problem optimizing a fleet of quadcopters for package delivery.
% We demonstrated that our approach outperforms the previous state-of-the-art by an order of magnitude, and showcased our CP subsystem technique on a larger, task-oriented scheduling problem where a fleet of quadcopters are assigned to deliver packages.

Currently, as the solver iteratively finds solutions on the PF, it must reset each time so that it can add a constraint that restricts that previous solution in subsequent solves. In future work, we intend to further speed up the solver by dynamically adding constraints as the solver is running. We also intend to investigate techniques of determining adjacent solutions once a solution has been found.

%%%%%%%%%%%%%%%%%%%%%%%%%%%%%%%%%%%%%%%%%%%%%%%%%%%%%%%%%%%%%%%%%%%%%%%%%%%%%%%%

%%%%%%%%%%%%%%%%%%%%%%%%%%%%%%%%%%%%%%%%%%%%%%%%%%%%%%%%%%%%%%%%%%%%%%%%%%%%%%%%

%%%%%%%%%%%%%%%%%%%%%%%%%%%%%%%%%%%%%%%%%%%%%%%%%%%%%%%%%%%%%%%%%%%%%%%%%%%%%%%%
% \vspace{-2mm}
% \section*{ACKNOWLEDGMENT}
% \vspace{-2mm}
% \noindent This material is based on work supported by the National Science Foundation grants NSF\#1846340, NSF\#2054744, and the GRFP DGE\#2139899. Any opinions, findings, and conclusions or recommendations expressed in this material are those of the author(s) and do not necessarily reflect the views of the National Science Foundation.

%%%%%%%%%%%%%%%%%%%%%%%%%%%%%%%%%%%%%%%%%%%%%%%%%%%%%%%%%%%%%%%%%%%%%%%%%%%%%%%%
\vspace{-2mm}

\bibliographystyle{IEEEtran}
\balance
\bibliography{references}

% \section{APPENDIX}

\end{document}